\documentclass[manuscript,screen,nonacm]{acmart} 

\AtBeginDocument{%
  }

\setcopyright{none}

\usepackage{booktabs} 
\usepackage{multirow}
\usepackage{arydshln}
\usepackage{subcaption} 

\usepackage{tikz}

\begin{document}

\title{A Framework for Task-Agnostic Counterfactual Fairness in Transformers through Activation Steering}

\author{Joshua Wolff Anderson}
\orcid{0000-0003-1543-6491}
\correspondingauthor
\affiliation{%
  \institution{University of Pittsburgh}
  \city{Pittsburgh}
  \state{Pennsylvania}
  \country{USA}
}

\author{Yanshan Wang}
\orcid{0000-0003-4433-7839}
\affiliation{%
  \institution{University of Pittsburgh}
  \city{Pittsburgh}
  \state{Pennsylvania}
  \country{USA}
}

\author{Shyam Visweswaran}
\orcid{0000-0002-2079-8684}
\affiliation{%
  \institution{University of Pittsburgh}
  \city{Pittsburgh}
  \state{Pennsylvania}
  \country{USA}
}

\renewcommand{\shortauthors}{Anderson et al.}

\begin{abstract}
Bias in transformer-based models is a growing concern as these architectures become more prevalent in automated decision-making systems, such as prediction models for hiring, lending, and clinical treatment recommendation. Mechanistic interpretability (MI) is a growing body of work that measures meaningful causal effects of activations on model output by identifying human-interpretable features encoded in model activations. When such features correspond to sensitive attributes, they provide a direct way to quantify the bias those attributes introduce and to causally intervene to mitigate it. However, counterfactual fairness (CF) methods have not been systematically adapted to this setting of intervening on interpretable mechanisms in transformers. We propose a framework, Mechanistic Interpretability for Counterfactual Fairness (MICF), that utilizes MI tools to reformulate CF constraints in terms of transformer activations, unifying former approaches into a single pipeline for fairness analysis in transformers. MICF does not require a prior causal model or specific transformer architecture. We evaluate MICF on real-world tabular data and show that CF can be improved across different prediction tasks.
\end{abstract}

\maketitle

\section{Introduction}

As machine learning (ML) models increasingly inform high-stakes decisions, from criminal risk assessment to clinical decision-making, demand for greater transparency in ML systems has grown substantially; transparency refers to the degree to which a model's internal computations are human-understandable \cite{lipton2018mythos}. Even prior to the rise of large language models (LLM), transformer architectures have increasingly been used for high-stakes decision-making such as clinical prediction \cite{mohamed2025advancing}, legal judgment prediction \cite{siino2025exploring}, hiring decisions \cite{tripathi2026mapping}, as well as in broader domains such as education, finance, marketing, and entertainment \cite{ofori2025explainable, annepaka2025large}. However, the scale and complexity of transformer models make their internal reasoning hard to inspect, limiting accountability for whether their decisions are reliable and equitable. As these models take on consequential roles, stakeholders have increasingly called for greater understanding and justification of their outputs, driving demand for transparency \cite{chen2024survey, tam2024framework, de2024artificial}. Two active areas concerning model transparency are interpretability and fairness. Interpretability methods aim to visualize or explain how a model maps input to output in a way understandable to humans, while fairness methods seek to ensure that its predictions do not encode bias \cite{kusner2017counterfactual, bereska2024mechanistic}. Yet, fairness and interpretability are typically pursued separately \cite{anderson2026effect}, and work at their intersection remains understudied.

Mechanistic interpretability (MI) is an interpretability framework that reverse-engineers neural networks' internal computations to identify components that causally mediate specific model behaviors \cite{bereska2024mechanistic}. MI differs from traditional post-hoc explanation methods like SHAP \cite{lundberg2017unified}, which approximate model behavior from the outside rather than examining its internal structure. MI enables inspection through mechanisms related to sensitive attributes that causally influence model predictions at the level of a model's internal representations, providing a principled basis for verifying fairness properties beyond aggregate output statistics.

Counterfactual fairness (CF) provides a complementary, causally grounded notion of algorithmic fairness: a prediction is fair for an individual if it would remain unchanged in a counterfactual world where they belonged to a different group \cite{kusner2017counterfactual}. Unlike group fairness metrics, CF respects the causal structure of sensitive attributes with respect to the outcome. A well-known limitation, however, is that traditional CF requires explicit structural causal models to measure causal effects, which are often non-identifiable in the presence of complex real-world confounding \cite{peters2012identifiability}.  

A unified framework is needed to explicitly specify and ground modern counterfactual generation methods, identify the causal structure of model internals, and, in principle, apply across transformer architectures and modalities to achieve CF. In this work, we propose Mechanistic Interpretability for Counterfactual Fairness (MICF), a framework that achieves these goals by connecting the normative concepts of CF with the mechanistic understanding of internal model computations. We leverage MI analysis of transformer internals to identify and intervene on the computational pathways through which sensitive attributes influence predictions.

\section{Background}

\subsection{Transformers}
We consider a standard $L$-layer transformer that maps an input sequence $\mathbf{x} = (x_1, \ldots, x_T) \in \mathcal{X}$ of $T$ tokens to an output logit vector $\hat{\mathbf{y}} \in \mathbb{R}^C$, where $C = |\mathcal{V}|$ is the vocabulary size \cite{vaswani2017attention}. Each token is embedded and combined with a positional encoding. At each layer $\ell$, an attention sublayer and then an MLP sublayer each add their output to the hidden state $\mathbf{h} \in \mathbb{R}^d$ (the residual stream). Normalization is omitted for clarity:

\begin{align}
  \mathbf{h}' &= \mathbf{h}^{(\ell-1)} + \mathrm{Attn}^{(\ell)}\!\left(\mathbf{h}^{(\ell-1)}\right), \\
  \mathbf{h}^{(\ell)} &= \mathbf{h}' + \mathrm{MLP}^{(\ell)}\!\left(\mathbf{h}'\right),
\end{align}

The output logits $\hat{\mathbf{y}}$ are obtained by applying a final normalization and an unembedding matrix $W_U \in \mathbb{R}^{C \times d}$ to the final residual stream at the last position, $\mathbf{z}^{(L)}_T$.

Because every sublayer writes its output additively into the residual stream, the stream at any layer $\ell$ is the embedding plus the sum of all earlier component outputs. This allows us to read, replace, or add a vector to the activations at a chosen layer while leaving the rest of the computation unchanged.

\subsection{Mechanistic Interpretability}
\label{sec:background:circuits}
 
MI decomposes model behavior into circuits \cite{olah2020zoom}. A circuit is a minimal subset of model components $\mathcal{C} \subseteq \mathcal{G}$ comprising attention heads, MLP layers, or the residual stream edges between them that are causally sufficient to reproduce a target model behavior. Components outside $\mathcal{C}$ do not causally contribute to that behavior.
 
This framing is significant as it connects model components to causal responsibility for a behavior in model outputs. A component that is not in the circuit for a given prediction carries no causal weight on that prediction, regardless of its activations. Circuits can be measured at different resolutions of model components, such as neurons or layers \cite{bereska2024mechanistic}. In this work, we measure layerwise components.

The primary tool for this localization of attributable mechanisms is activation patching \cite{pearl2022direct, meng2022locating, heimersheim2024use}. Given a "clean" input $\mathbf{x}$ and a "corrupted" (or counterfactual) input $\mathbf{x}'$, activation patching measures the causal effect of a circuit on the output by replacing its activation from the corrupted run into the clean run and observing the resulting change in output. For example, the value of a sensitive feature $S$ is changed from $s$ to $s'$. Both runs are cached, allowing for the replacement (patching) of corrupted activations at a given layer.

A patching experiment at component $(\ell, k)$ then runs a third forward pass (the patched run), which is identical to the clean run except that the activation of component $(\ell, k)$ is replaced with its value from the corrupted run to act as a causal intervention on the activation of that particular component:
\begin{equation}
  \mathbf{a}^{(\ell,k)} \;\leftarrow\; \mathbf{a}'^{(\ell,k)}
  \label{eq:patch_op}
\end{equation}
The causal effect of the intervention on model output is the resulting change in an output metric
$\mathcal{M}$ (e.g., the logit difference between two classes):
\begin{equation}
  \Delta^{(\ell,k)}
  \;=\;
  \mathcal{M}\!\left(f^{\mathrm{patch}(\ell,k)}(\mathbf{x},\,\mathbf{x}')\right)
  \;-\;
  \mathcal{M}\!\left(f(\mathbf{x})\right).
  \label{eq:patch_effect}
\end{equation}
If $\Delta^{(\ell,k)} \approx 0$, patching that component has no effect; it is not on a causal path between the corrupted input and the output difference. If $\Delta^{(\ell,k)}$ is large, the component causally mediates the difference. By sweeping over all components $(\ell, k)$ and measuring each $\Delta^{(\ell,k)}$, we can observe which layers or heads are causally responsible for a given behavior.

Activation patching can be interpreted as an instance of Pearl's \emph{do}-calculus \cite{pearl1994probabilistic}: replacing $\mathbf{a}^{(\ell,k)}$ with its corrupted value corresponds to an intervention $\mathrm{do}\!\left(\mathbf{a}^{(\ell,k)} = \mathbf{a}'^{(\ell,k)}\right)$ on the computational graph of the model. This distinguishes patching from probing or linear classification, which measure correlations between activations and attributes but do not identify causal mediation \cite{bereska2024mechanistic}. Formally, the goal of CF is to measure an intervention for a sensitive feature $S$:

\begin{equation}
  P\!\left(\hat{Y} \mid \mathrm{do}(S = s^\prime)\right)
  \;\neq\;
  P\!\left(\hat{Y} \mid S = s\right)
  \label{eq:do_vs_obs}
\end{equation}

\noindent MI methods can do this by finding model activations that represent $S$. A key concept underlying much of MI is the \emph{linear representation hypothesis} \cite{park2023linear}, which states that human-interpretable features can be represented linearly as directions in the model activation space. A feature $f$ is said to be linearly represented if there exists a direction $\hat{\mathbf{v}} \in \mathbb{R}^d$, $\|\hat{\mathbf{v}}\|=1$, such that the presence of $f$ in an input corresponds to a large inner product $\langle \mathbf{x}^{(\ell)}, \hat{\mathbf{v}} \rangle$ at some layer $\ell$.
 
Representations may be harder to isolate cleanly under superposition \cite{elhage2022toy}, where models encode more features than they have dimensions using near-orthogonal directions. As a result, $S$ could be partially distributed across multiple near-orthogonal directions rather than isolated to a single one. This does not invalidate direction-based measurements of $S$. Still, it means the true direction of an attribute's vector might not be exactly what is measured. Disentangling superposed features remains an active area of interpretability research \cite{gantla2025exploring}. We intervene on whole-layer activations to capture the bulk of the sensitive attribute’s signal, at the cost of specificity: the direction we identify may also carry correlated or unrelated information, and finer-grained (e.g., neuron-level) attribution is complicated by feature entanglement. We leave a more complete treatment of this to future work, as disentanglement methods mature.

\subsection{Counterfactual Generation}

A key challenge in enforcing counterfactual fairness is the need to generate plausible counterfactual inputs: samples that reflect what an individual's feature vector would look like under a different value of the sensitive feature $S$. Under counterfactual fairness \cite{kusner2017counterfactual}, a model's prediction for an individual should remain unchanged when $S$ is intervened upon, so the quality of the counterfactuals directly determines how well this constraint can be enforced.

Our approach builds on the notion of nearest counterfactuals, which seeks the feature configuration that produces a desired model outcome while maintaining a minimal distance from the individual's original feature vector \cite{karimi2020model}. Nearest counterfactuals provide informative contrasts to factual samples for explaining models, such as in MI methods. Generating them in our setting requires knowledge of how features depend on one another, which is typically achieved with a structural causal model. Recent work suggests that Large Language Models (LLMs) implicitly encode causal reasoning because they are trained on large, diverse corpora \cite{narendra2025causal, youssef2024llms, mishra2024llm}, making them a candidate to approximate this domain-specific knowledge.

Under the assumption that LLMs encode sufficient domain-specific knowledge, the resulting samples must satisfy four properties: proximal (small deviation from the input), plausible (adherence to real-world constraints), causally consistent (respect for structural dependencies), and actionable (changes that can be practically implemented) \cite{narendra2025causal, laugel2019dangers}. Failing these requirements can lead to unrealistic feature values or violations of causal dependencies.

We adopt the agentic counterfactual generation framework of \citet{narendra2025causal}, in which candidate counterfactuals are iteratively refined through self-consistency checks. Each iteration uses chain-of-thought style prompts that guide the LLM toward the four properties above. The validity of the counterfactuals can be measured using the Parent-Child Causal Violation Rate (PC-CVR) from \citet{narendra2025causal}, which measures the fraction of counterfactuals that violate known causal structural dependencies. PC-CVR is defined as:

\begin{equation}
    \label{eq:pc-cvr}
    \text{PC-CVR} = \frac{1}{N} \sum_{i=1}^N I\left( \bigvee_{(P_j,C_j) \in \mathcal{C}} \delta_{{i,j}}(P_i, C_i) \right)
\end{equation}

\noindent where $I(\cdot)$ is the indicator function, $\bigvee(\cdot)$ is logical OR over all parent-child pairs, $N$ is the number of samples being evaluated, and $\delta(\cdot)$ is a function that tests if a parent $P$ and child $C$ pair violate known causal structural dependencies \cite{narendra2025causal}. Because counterfactual generation is delegated to a generative model, the framework is modality-agnostic and applies wherever a suitable generative model exists.

Counterfactual samples from this process mirror a broader concept of causal effect estimation from observational data: true counterfactual outcomes are unobservable, so recovered effects using approximate counterfactuals are best understood as approximations that move closer to the causal truth rather than recovering it exactly \cite{pearl2010introduction, hernan2020whatif}. Latent confounding may leave residual bias in these effect estimates. Because the true causal model generally cannot be validated, prior work has argued for enforcing counterfactual fairness approximately, and across plausible causal models, instead of requiring an exactly specified one \citep{russell2017worlds}. We constrain using plausible approximate causal effects, which, under the assumption that LLMs encode such knowledge, can still reduce a causal source of bias in model outputs. The effectiveness of the bias mitigation is dependent on the assumed causal model that generates the counterfactuals.

\section{Related Works}

\citet{singla2021using} propose a framework for deriving global explanations of deep learning classifiers using causal mediation analysis. They identify hidden units correlated with clinically defined concepts via sparse logistic regression, then verify causal relevance by
treating each set of concept units as a mediator in the pathway from input to output. Using Pearl's direct and indirect effects via the do-operator, they rank concepts by the degree to which counterfactual perturbations propagate through those units to change the
prediction, rather than bypassing them. While this work was published in parallel to MI literature and does not use that framing, it applies the same core intuition that internal units should be evaluated by their causal influence on model outputs, not merely their correlation with a variable of interest in imaging data. 

Recent work on prior-data fitted networks (PFNs) and in-context learning (ICL) offers a promising avenue for embedding causal reasoning directly into model training for tabular transformers. TabPFN \cite{hollmann2025accurate} is a transformer-based foundation model trained over synthetic datasets sampled from structural causal models, enabling generalization across classification tasks via a single forward pass. Because TabPFN's training prior encodes a broad space of causal structures, FairPFN \cite{robertson2024fairpfn} proposes leveraging this to enforce counterfactual fairness during prior fitting, eliminating the need for an explicit causal model at inference time. However, PFN-based models degrade under covariate shift \cite{liu2026causal}, and the fairness optimization in FairPFN remains specific to the PFN architecture, precluding generalization to transformer architectures trained on standard supervised objectives.

Dong and Quan \cite{dong2026efficient} propose the Feature Correlation Transformer (FCorrTransformer), an attention-light architecture for tabular data that enforces counterfactual fairness through Counterfactual Attention Regularization (CAR). FCorrTransformer modifies the standard scaled dot-product attention by removing query, key, and value projections entirely, so that the attention matrix reduces to pairwise scalar similarities between one-dimensional feature embeddings, admitting a direct statistical interpretation as a feature dependency structure. CAR enforces counterfactual fairness by penalizing the discrepancy between attention weights computed on the original input and those computed on counterfactual permutations of the sensitive feature $S$, with the full training objective $\mathcal{L} = \mathcal{L}_{\mathrm{perf}} + \lambda \mathcal{L}_{\mathrm{CAR}}$. Empirical results on the Bank Account Fraud and InsurTech datasets demonstrate strong counterfactual fairness with minimal performance loss. However, the fairness enforcement depends on regularizing a classifier and does not generalize for a foundation model across different tasks. The authors also explicitly acknowledge that CAR does not establish formal causal guarantees regarding the relationship between sensitive attributes and predictions.

Recent work has begun to explore inference-time activation interventions as an alternative to training-time debiasing in transformer-based models. FairSteer \cite{li2025fairsteer} proposes an LLM debiasing framework that trains a linear classifier on intermediate-layer activations to detect bias signatures, computes debiasing steering vectors as the mean difference between biased and unbiased activation clusters, and applies these vectors dynamically at inference time. While FairSteer demonstrates that fairness-related features are linearly separable in the transformer activation space, its identification of bias-relevant components is correlational rather than causal: the linear classifier learns directions associated with bias labels rather than using activation patching to isolate the causal influence of sensitive attributes on model predictions.

Across these works, a range of useful methodologies has emerged for analyzing and debiasing neural models: Singla et al. \cite{singla2021using} demonstrate that causal mediation analysis can identify internal units that genuinely drive model decisions in image classifiers, FairSteer \cite{li2025fairsteer} establishes the feasibility of activation-space intervention at inference time for language models, FCorrTransformer \cite{dong2026efficient} shows that architectural design can enforce interpretable fairness constraints in tabular transformers, and FairPFN \cite{robertson2024fairpfn} demonstrates that transformers can internalize causal fairness objectives during training on tabular data. Notably, these works span different data modalities (medical images, natural language, and tabular data) and were developed independently, without a shared theoretical foundation; they vary in assumptions, intervention sites, and fairness guarantees in ways that make them difficult to compare or systematically extend. Although they all work towards a common goal of CF in deep learning using interpretable methods, MICF unifies these works through a standardized framework: counterfactual generation, identification of causal bias through interventions using MI, and CF through activation steering. 

\section{Methods}

\begin{figure}[t]
    \centering
    \includegraphics[width=\linewidth]{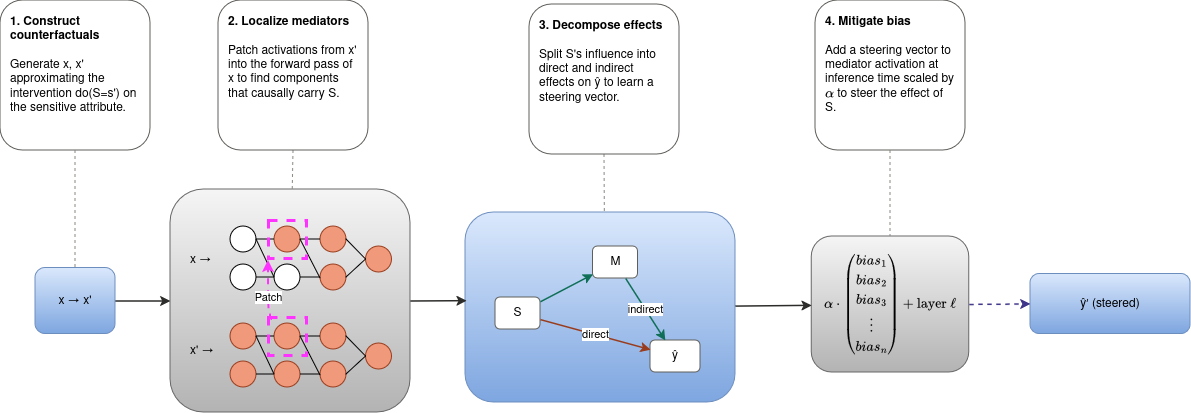}
    \caption{Overview of the MICF framework.}
    \label{fig:micf-framework}
    \Description{Visual representation of the steps in the MICF framework.}
\end{figure}

The MICF framework instantiates a universal pipeline for CF analysis in transformers as four concrete stages operating directly on activations: (i) construct counterfactual inputs that approximate the intervention $do(S{=}s')$; (ii) localize the components that causally mediate the influence of the sensitive feature on the prediction; (iii) decompose that influence into direct and indirect causal effects; and (iv) mitigate bias at inference time through activation steering. Because each step generalizes across data modalities and transformer architectures, the same procedure remains unchanged regardless of the problem. Figure \ref{fig:micf-framework} summarizes each step.

We instantiate the framework on a single-estimator classifier, TabPFN v2, using ICL to provide our data as context \cite{hollmann2025accurate}. By default, input features for TabPFN are preprocessed (removal of empty features, NaN handling, normalization) and linearly projected to a 192-dimensional embedding, while class labels are separately encoded and projected to the same dimension. The model architecture is a stack of 12 identical layers, with a transformer encoder applying two-way self-attention (first across features, then across rows) followed by a two-layer  LP $(192 \rightarrow 768 \rightarrow 192)$ and layer normalization. A final MLP decoder $(192 \rightarrow 768 \rightarrow 10$ with GELU activation) maps the representation of each test sample to class logits for predictions. In our analysis, activations are taken from the output of the $\ell$-th transformer layer, i.e., the residual stream after the encoder's MLP and normalization.

\subsection{Counterfactual Generation and Fairness}

For each dataset, we generate two sets of factual-counterfactual pairs under the intervention $do(S = s^\prime)$. The first is \emph{flip}, in which we change only the binary sensitive feature $S$ to its alternative value and leave all other features invariant. This baseline ignores causal propagation, so features that depend on $S$ retain their factual values. The second set is \emph{LLM}, which feeds an LLM the factual record and the intervention and outputs any modifications to features it judges to be causally affected by $S$. 

Our implementation generates the LLM-based counterfactuals using the agentic generation loop of \citet{narendra2025causal}. We use \texttt{llama3:8b-instruct-q8\_0}, an 8-billion-parameter instruction-tuned LLM with 8-bit quantization, served locally through Ollama (v0.30.10) on a single NVIDIA RTX 4070 GPU (12\,GB VRAM). Running the model locally avoids sending sensitive electronic health records (EHR) included in our experiments to external services. The LLM is given the following system prompt:

\begin{quote}
\textit{``You are a counterfactual explanation agent for a tabular classifier. You propose minimal, realistic, actionable feature changes that reflect real-world causal dependencies between features. Reply with JSON only.''}
\end{quote}

\noindent For each instance $x$ with $S \neq s^\prime$, the user prompt contains the feature values as JSON, a schema (numeric ranges, allowed categorical values, integer-only counts, etc.), and the question of what the record would look like had $do(S=s^\prime)$ occurred. The prompt also instructs the model to propagate the change only to features with a plausible causal pathway from $S$, to avoid stereotypical effects, and to leave all other features identical. The model returns a counterfactual as a JSON object, which is parsed and validated against the schema. Candidates are refined for up to 5 iterations. Instances for which no valid counterfactual is produced are counted as failures and excluded. We sample a minimum of 300 instances per dataset with a fixed seed, and use a temperature of 0.7. No causal rules are provided during generation to require the LLM to rely on its own implicit knowledge.

We measure causal consistency with the PC-CVR metric defined by \citet{narendra2025causal}. In the original formulation, the parent-child dependencies $\mathcal{C}$ are specified as hand-written, rule-based constraints. This requires manual, domain-specific effort for each dataset. To make our pipeline fully automated, we instead estimate a causal structure from the data using the LiNGAM causal discovery algorithm \cite{shimizu2006linear}. The model is fit on a given dataset, and the resulting directed edges form the set $\mathcal{C}$. We measure PC-CVR using (\ref{eq:pc-cvr}) with the edges in $\mathcal{C}$.

Because $\mathcal{C}$ is learned rather than specified by experts, PC-CVR measures consistency with the causal structure recoverable from the observed data, not with a ground-truth causal graph. LiNGAM assumes linear relationships, non-Gaussian noise, and no hidden confounders, so its output may be unreliable when significant latent confounding is present. We therefore treat PC-CVR as a practical, automated proxy for causal consistency and do not interpret it as evidence of exact causal effects.

We quantify a fairness metric $cf_\text{score}$ on the model output as the distance between the classification model's predicted probabilities for the factual and counterfactual samples. This is a common method for measuring counterfactual fairness \cite{plevcko2024causal,ma2023learning,anderson2025algorithmic}:
\begin{equation}
  cf_\text{score} \;=\; \frac{1}{n}\sum_x
    \left| P(\hat{Y}\mid x') - P(\hat{Y}\mid x) \right|
\end{equation}
This is the average magnitude of the total causal effect (TE) of $S$ on $\hat{Y}$ for across individuals. A value of $cf_\text{score}{=}0$ implies perfect counterfactual fairness, and the prediction is invariant to $S$. All localization, mediation, and mitigation results are reported relative to this score.

\subsection{Localizing Sensitive-Attribute Mechanisms}

To identify which layer carries the strongest signal in $S$'s influence, we perform ablation-based attribution patching. For each layer $\ell$, we ablate the activations by zeroing the corresponding block of its output projection, $W^{(\ell)}_{\text{out}} \leftarrow \mathbf{0}$, which removes that layer's contribution to the residual stream while leaving every other component intact. We then recompute $cf_\text{score}$ on the ablated model, $cf_{\text{abl}}^{(\ell)}$, and measure the fraction of bias that the head accounts for,
\begin{equation}
  r^{(\ell)} \;=\; 1 - \frac{cf_{\text{abl}}^{(\ell)}}{cf_{\text{base}}},
\end{equation}
where $cf_{\text{base}}$ is the unablated score. A large recovery $r^{(\ell)}$ means that disabling layer $\ell$ substantially reduces the causal effect of $S$ on the output, identifying that as part of the circuit through which the sensitive feature propagates. Sweeping all $\ell$ and ranking by $r^{(\ell)}$ yields a map of where bias from $S$ is computed, and pinpoints the target for activation steering.

\subsection{Causal Mediation Analysis}

Ablation helps localize bias but does not precisely measure the direction that activations of a given layer should steer to mitigate bias. To make this distinction, we perform a mediation analysis at each layer $\ell$, decomposing the total effect into a natural direct effect (NDE) and a natural indirect effect (NIE). We first cache the query-position activations of layer $\ell$ under both the factual input $x$ and the counterfactual input $x'$. The three effects are then obtained from patched forward passes as:
\begin{align}
    \text{TE}  &= P(\hat{Y} | x') - P(\hat{Y} | x)\\
  \text{NDE} &= P\!\left(\hat{Y}\mid x',\, a^{(\ell)} {\leftarrow} a^{(\ell)}_{\text{clean}}\right) - P(\hat{Y}\mid x), \\
  \text{NIE} &= P\!\left(\hat{Y}\mid x,\, a^{(\ell)} {\leftarrow} a^{(\ell)}_{\text{corrupted}}\right) - P(\hat{Y}\mid x).
\end{align}
We run the factual input while patching the layer's activation to its counterfactual value. This blocks the layer's contribution, so its contribution to the residual stream is from the counterfactual activations. These definitions of TE, NDE, and NIE follow standard causal mediation analysis \cite{pearl2022direct} and its application to causal fairness \cite{plevcko2024causal}. Our interventions target minimizing TE, but we also report NDE and NIE to show that mitigations can target these decomposed pathways.

\subsection{Activation Steering for Bias Mitigation}

We mitigate the influence of $S$ by steering its related activations at inference time, without retraining or modifying the architecture. Exploiting the linear representation hypothesis, we estimate the direction along which the sensitive feature is encoded as the difference between the target layer's activations on the factual and counterfactual runs. This steering vector $\mathbf{v}_s$ is a real-valued vector that specifies how to shift activations away from the biased direction in the activation space. We additively intervene on the activations to push representations away from the direction associated with the sensitive feature's effect on model predictions:

\begin{equation}
  \mathbf{v}_s \;=\;
    a^{(\ell)}_{\text{clean}} -
    a^{(\ell)}_{\text{corrupted}}
\end{equation}
Additionally, we normalize $\mathbf{v}_s$:

\begin{equation}
    \mathbf{v}_\text{norm} = \frac{{\mathbf{{v}}}_s}{\lVert {\mathbf{v}}_s \rVert_2}
\end{equation}

\noindent then scale by coefficient $\alpha$ that controls the strength of the intervention:
\begin{equation}
  a^{(\ell)} \;\leftarrow\; a^{(\ell)} + \alpha\,\mathbf{v}_s
\end{equation}
 The strength $\alpha$ interpolates between the unmodified model ($\alpha{=}0$) and progressively stronger correction to steer the activation vectors away from the sensitive direction, allowing the effect of $S$ to be driven toward zero.

Because steering trades fairness against predictive performance, we select $\alpha$ by testing a range of values and recomputing both the effects and task accuracy on the steered model. This traces the bias-accuracy trade-off well known in the fairness literature \cite{plecko2025fairness}. We identify an $\alpha$ value that reduces the effect of $S$ while preserving sufficient accuracy. We bootstrap 1000 resamples with replacement for each candidate $\alpha$, and report $95\%$ confidence intervals for both quantities. 

\subsection{Statistical Testing}

To test whether steering changes a metric relative to the unsteered model $(\alpha=0)$, we compare the selected steering strength $(\alpha=75)$ using a paired nonparametric bootstrap over evaluation examples. We draw $N$ examples with replacement, where $N$ is the dataset size, keeping each factual input consistent with its counterfactual. Both $\alpha$ values are evaluated on the same resample. For each metric $m \in \{TE, NDE, NIE, Accuracy\}$, this yields a bootstrap distribution of the difference $\Delta_m = m_{\alpha = 75} - m_{\alpha = 0}$. We focus our report on the mean of this distribution for Accuracy and TE and a 95\% confidence interval (CI). Formally, we can state our statistical hypothesis as the following:

\begin{align*}
    H_0 &: \Delta_m = 0\\
    H_1 &: \Delta_m \neq 0
\end{align*}

\begin{figure*}[ht!]
\centering
\captionsetup[subfigure]{font=footnotesize, labelfont=footnotesize, textfont=footnotesize}

\renewcommand{\arraystretch}{0.75}
\newcommand{\pw}{0.42\textwidth}
\setlength{\tabcolsep}{2pt}

\newcommand{\steerpanel}[2]{%
  \begin{subfigure}{\linewidth}
    \includegraphics[trim={0 0 0 1.2cm}, clip, width=\linewidth]{figures/#1}
    \caption{#2}
    \Description{Line plot of accuracy and the direct, indirect, and total counterfactual fairness effects against steering strength alpha.}
  \end{subfigure}}

\newcommand{\cell}[2]{\begin{minipage}[c]{\pw}\steerpanel{#1}{#2}\end{minipage}}

\newcommand{\rowsep}{\\\noalign{\vskip 0.5em}\cdashline{2-3}\noalign{\vskip 0.5em}}

\newcommand{\labelgap}{0.5em}  
\newcommand{\rulegap}{1.5em}   

\newcommand{\rowlabel}[1]{%
  \makebox[1.2em][c]{\rotatebox[origin=c]{90}{\small\textbf{#1}}}%
  \hspace{\labelgap}%
  \vrule width 0.4pt%
  \hspace{\rulegap}}

\begin{tabular}{@{}c@{}c@{\hspace{2em}}c@{}}

\rowlabel{$\text{UTICalc}_{flip}$} &
  \cell{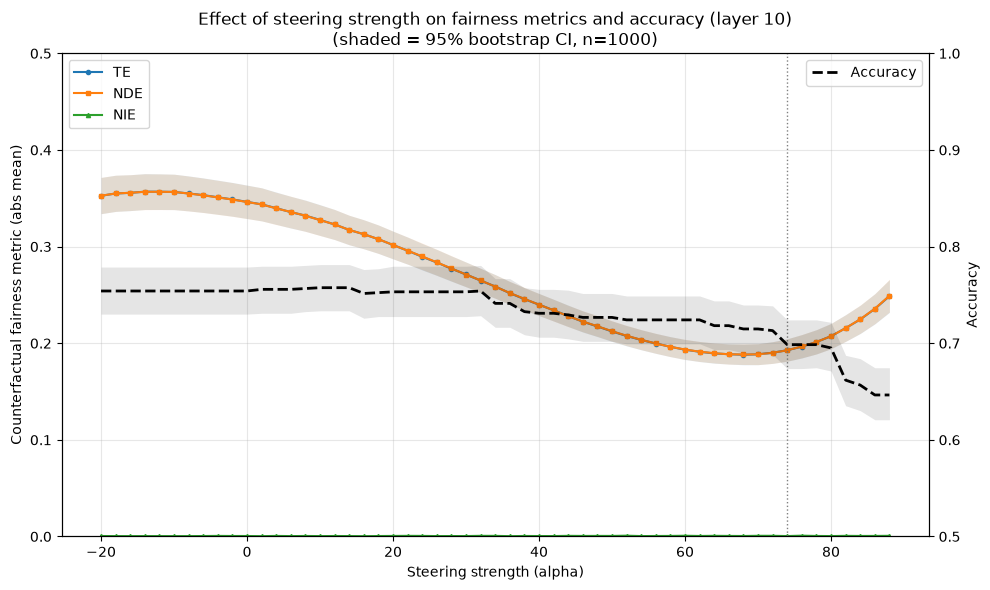}{$\text{UTICalc}_{flip}$} &   
  \cell{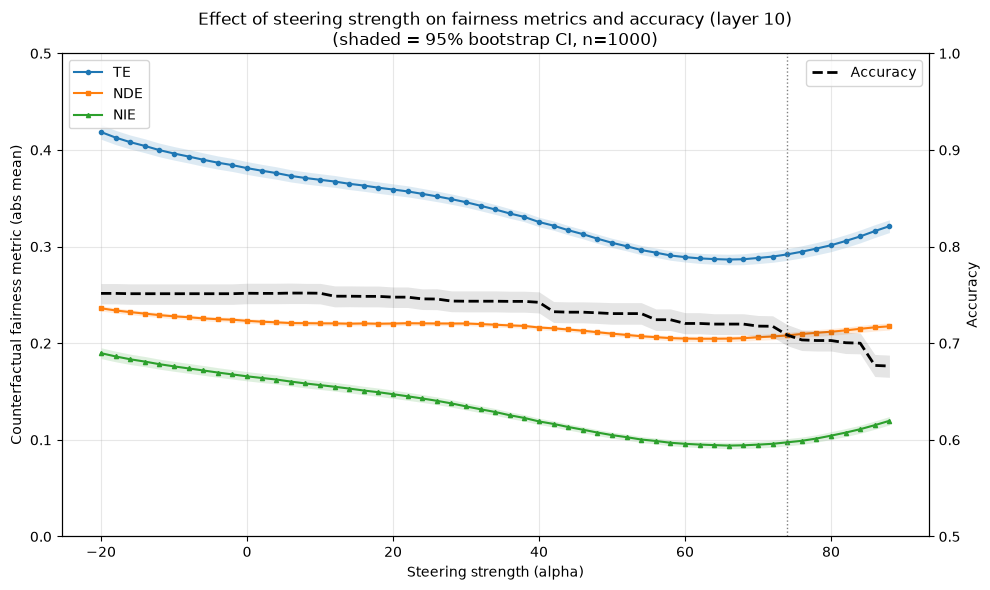}{$\text{COMPAS}_{flip}$}       
\rowsep
\rowlabel{$\text{COMPAS}_{flip}$} &
  \cell{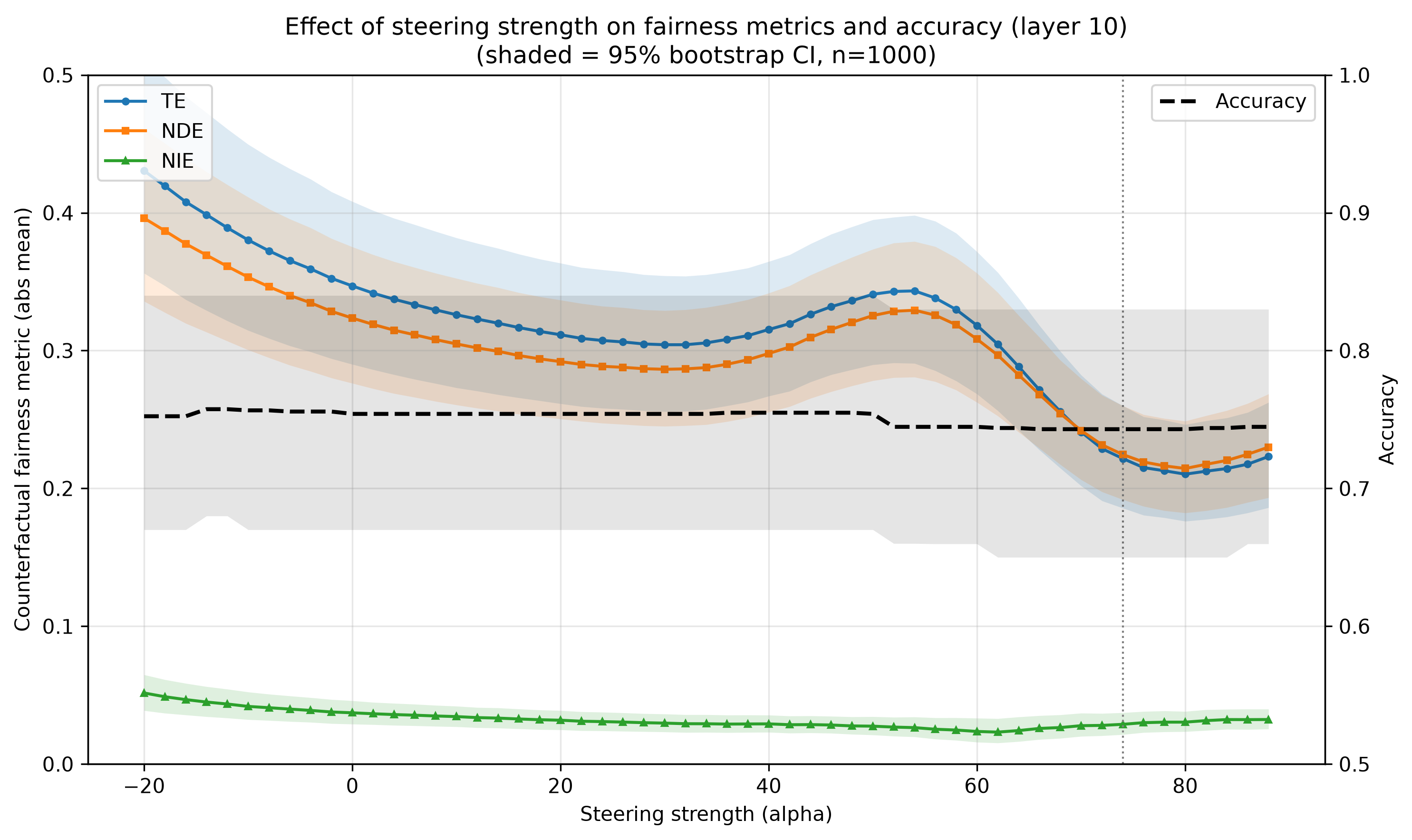}{$\text{UTICalc}_{flip}$} &   
  \cell{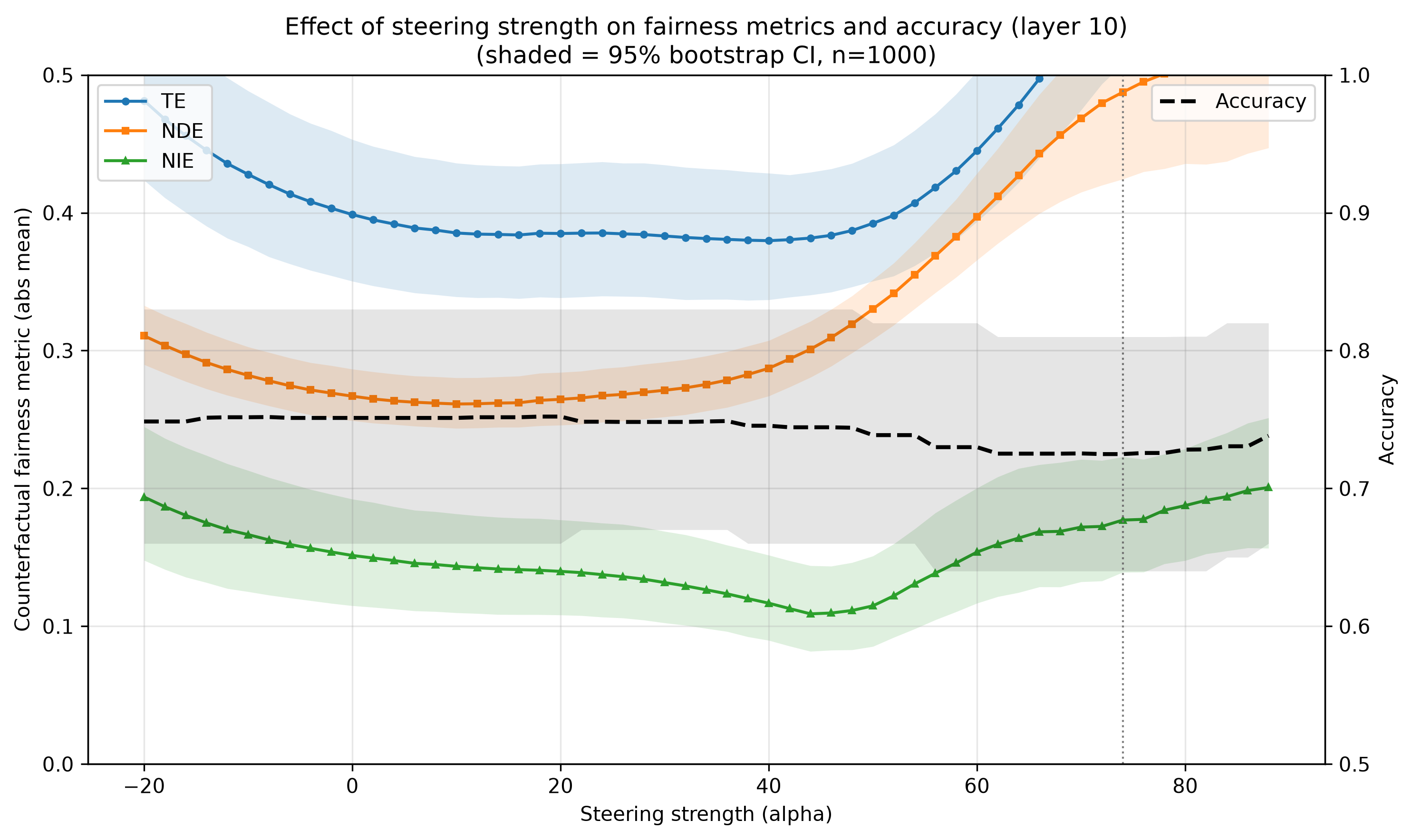}{$\text{COMPAS}_{flip}$}   
\rowsep

\rowlabel{$\text{UTICalc}_{LLM}$} &
  \cell{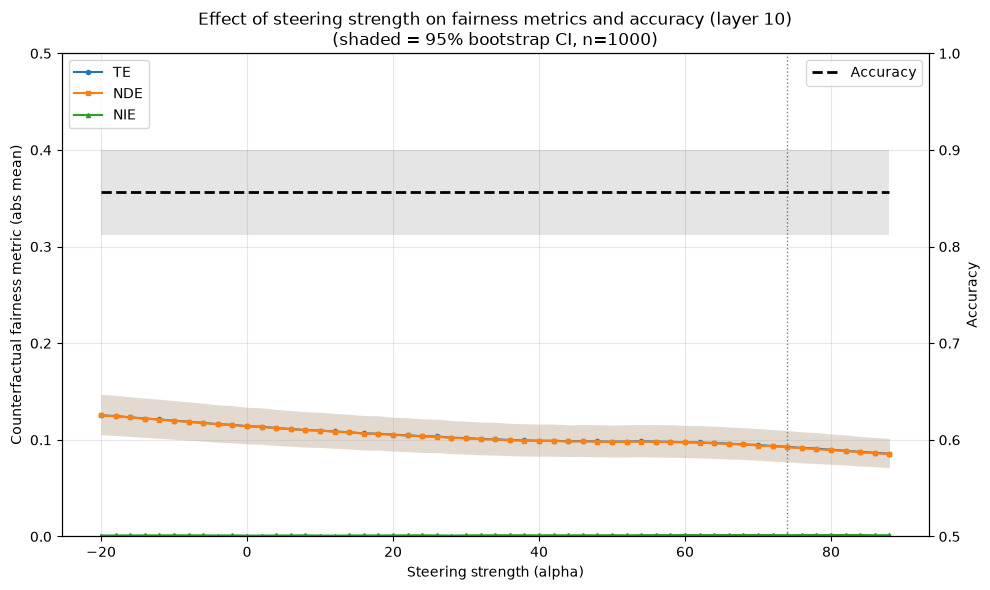}{$\text{UTICalc}_{LLM}$} & 
  \cell{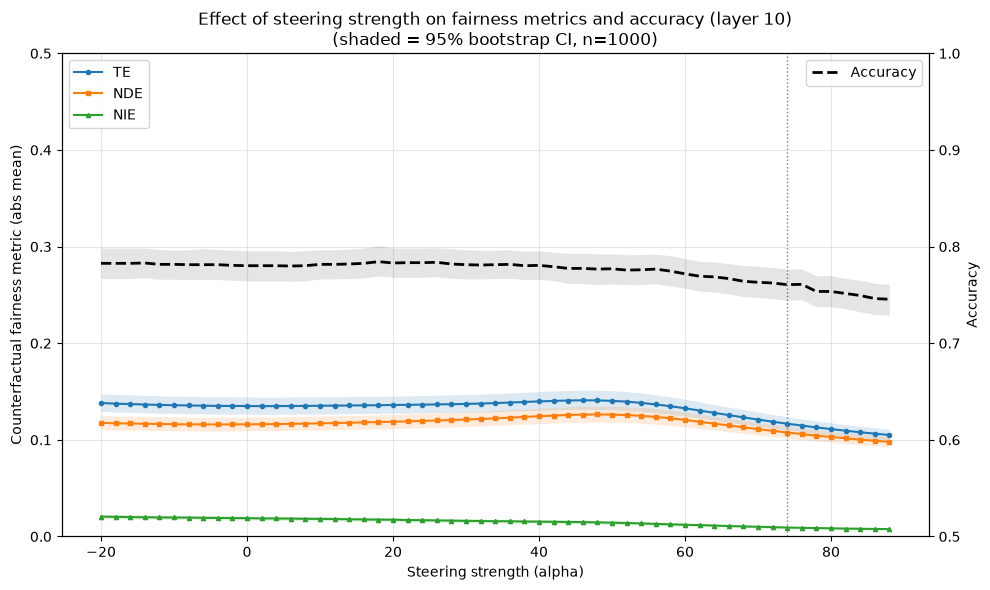}{$\text{COMPAS}_{LLM}$}     
\rowsep
\rowlabel{$\text{COMPAS}_{LLM}$} &
  \cell{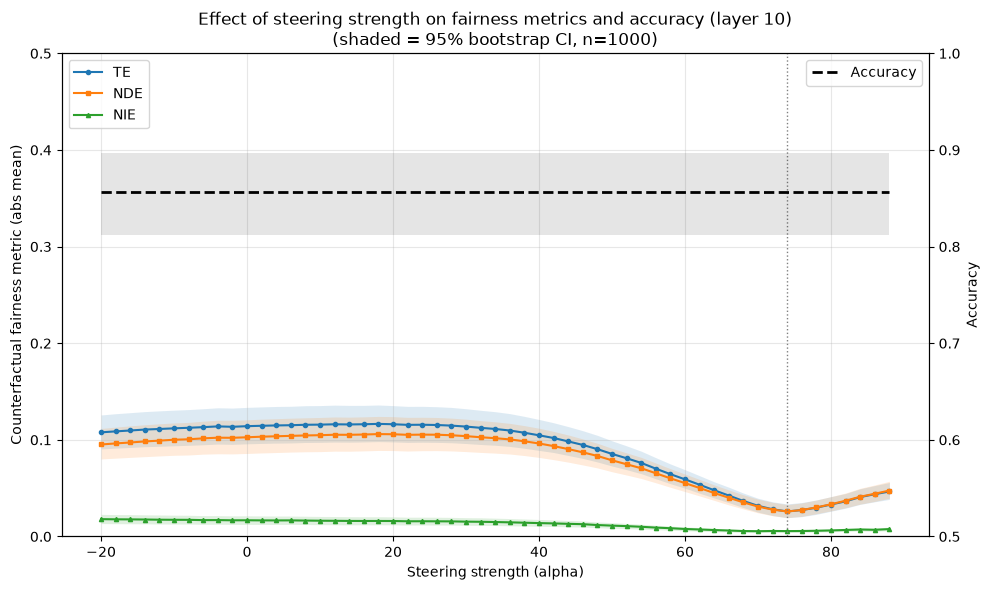}{$\text{UTICalc}_{LLM}$} & 
  \cell{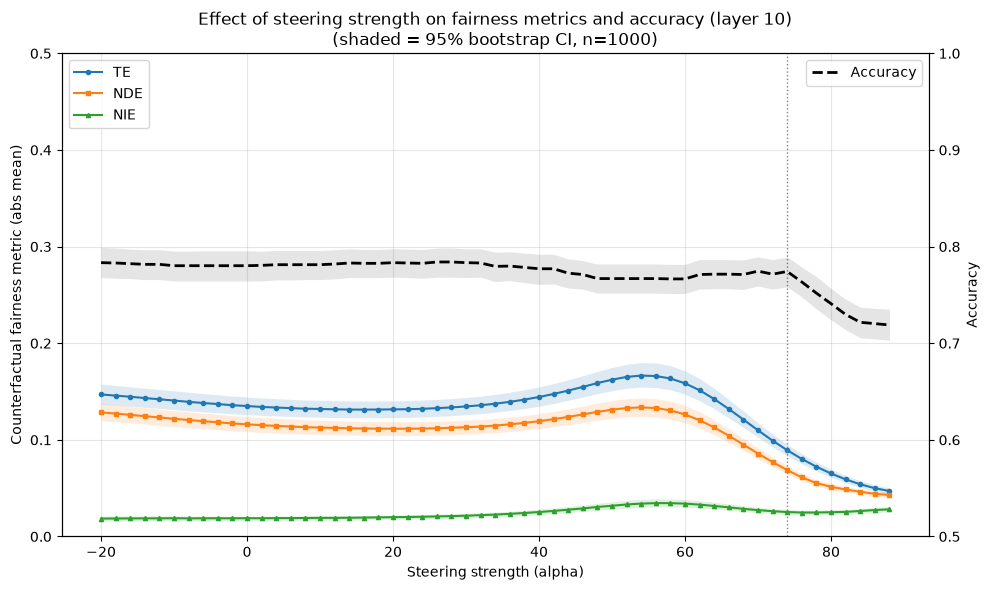}{$\text{COMPAS}_{LLM}$}     
\\
\end{tabular}

\caption{Counterfactual fairness results under activation steering. Rows are grouped by the counterfactual data used to learn the steering vector: flip (top two rows) and LLM (bottom two rows), showing results for UTICalc and COMPAS, respectively. Each panel plots accuracy and counterfactual fairness metrics (y-axis) against the steering strength $\alpha$ (x-axis). Black lines show accuracy; orange, green, and blue lines show the direct, indirect, and total effects, respectively.}
\Description{Grid of line plots of accuracy and the direct, indirect, and total counterfactual fairness effects against steering strength alpha.}
\label{fig:steer_mitigation}
\end{figure*}

\noindent The same bootstrap distributions are used to draw the shaded 95\% percentile bands in Figure ~\ref{fig:steer_mitigation}. 

\subsection{Datasets}

We evaluate on two real-world tabular benchmarks with distinct domains and a binary race variable as the sensitive feature. The first is UTICalc, a clinical urinary tract infection risk model shown to have significant racial bias \cite{shaikh2022reassessment,anderson2024measuring}. Because collecting a urine sample from children under 2 years of age is difficult, the UTICalc model estimates the probability of UTI using clinical features readily obtained in the emergency department \cite{shaikh2018development}. The tool helps clinicians identify children most at risk of UTI to decide whether catheterization is warranted for timely urine testing. The dataset consists of 1,686 children and seven binary clinical and demographic predictors: age ($\geq$12 months vs. $<$12 months), race (non-Black vs.\ Black), sex (female or uncircumcised male vs.\ circumcised male), alternate fever source (no other source of fever vs. other source of fever), maximum temperature ($\geq$39\textdegree C vs.\ $<$39\textdegree C), history of UTI (yes vs.\ no), and fever duration ($\geq$48 hours vs.\ $<$48 hours).

The second dataset is Northpointe Inc.’s Correctional Offender Management Profiling for Alternative Sanctions (COMPAS) dataset, which contains records of defendants from Broward County, Florida. It is widely used to study algorithmic fairness in criminal justice since ProPublica published a report on the model's discrimination against African-American defendants \cite{angwin2016machine}. The prediction task is to classify whether a defendant will reoffend within two years of being scored (two-year recidivism). We use six features from the dataset: sex (male vs.\ female), age ($<$25, 25--45, or $>$45 years), race (African-American vs. Other), the number of prior adult offenses (count), the number of prior juvenile misdemeanors (count), and juvenile felonies (count). The first three features are categorical, and the last three are non-negative integer counts. We modified the original race values that included more granular categories (e.g. Caucasian, Hispanic, Asian-American, etc) to a binary value (African-American vs. Other) to match the fairness problem of UTICalc. We use 6,172 defendants in our analysis.

We evaluate baseline predictive performance on the training data for both datasets with AUROC and accuracy, and baseline fairness with equalized odds. Equalized odds is a group fairness metric that requires that the model's predictions be independent of the sensitive feature $S$ conditional on the true outcome $Y$, such that $P(\hat{Y}=1 \mid S=s, Y=y)$ should be equal across values of $s$ for both $y \in \{0, 1\}$. In practice, we report the equalized odds difference, defined as the larger of the absolute differences in true positive rate (TPR) and false positive rate (FPR) between groups \cite{bird2020fairlearn}:
\begin{equation}
    \text{EO} = \max\left( \left|\text{TPR}_{s=0} - \text{TPR}_{s=1}\right|,\ \left|\text{FPR}_{s=0} - \text{FPR}_{s=1}\right| \right),
\end{equation}
where a value of 0 indicates perfect equalized odds. These metrics are computed using scikit-learnn (v1.8.0) and fairlearn (v0.14.0) packages for Python.

\begin{table}[t]
\centering
\caption{TabPFN performance of models trained on factual samples without any fairness constraint. EO is the equalized odds difference (lower is fairer).}
\label{tab:baseline}
\begin{tabular}{lccc}
\toprule
Dataset & AUROC $\uparrow$ & Accuracy $\uparrow$ & EO $\downarrow$ \\
\midrule
UTICalc (n=1171) & 0.809 & 0.755 & 0.171 \\
COMPAS (n=6172)  & 0.825 & 0.748 & 0.315 \\
\bottomrule
\end{tabular}
\Description{Table of TabPFN results for both datasets including AUROC, Accuracy and Equal Opportunity Difference.}
\end{table}

\section{Results}
Table~\ref{tab:baseline} reports baseline performance for the models trained without any fairness constraint. For UTICalc, TabPFN achieves an AUROC of 0.809 and an accuracy of 0.755, with an equalized odds difference of 0.171. For COMPAS, the model achieves a higher AUROC of 0.825 and a similar accuracy of 0.748, but a substantially larger equalized odds difference of 0.315. Predictive performance is therefore comparable across the two datasets, while the disparity in error rates between groups is nearly twice as large for COMPAS. This indicates that, before any intervention, the COMPAS model appears to have a stronger signal of racial bias than UTICalc. 

Activation patching recovered the largest share of the racial-bias signal at layers 9 and 10 $(r^{(9)}=1.3\%\ \text{and}\ r^{(10)}=1.6\%)$, whereas layers 0 and 11 recovered almost none $(r^{(0)}=0.01\%\ \text{and}\ r^{(11)}=1.2\times10^{-5}\%)$. All other layers ranged between 0.1\% and 0.4\%. Given that layer 10 yielded the highest bias recovery, we use its activations in all subsequent analyses.

When analyzing counterfactuals using the LLM method, we retain only counterfactuals that are consistent with no PC-CVR violations. After filtering, we retain 271 factual-counterfactual pairs for UTICalc (from 398 generated) and 2853 pairs for COMPAS (from 3,696 generated), corresponding to PC-CVR rates of 0.319 and 0.228, respectively. 

Figure~\ref{fig:steer_mitigation} shows how counterfactual fairness metrics and accuracy vary with steering strength $\alpha$, evaluated on factual--counterfactual prediction pairs. Each pair of subfigures corresponds to one steering vector, evaluated on UTICalc and COMPAS, respectively: (a, b) the vector learned from UTICalc$_{flip}$ samples, (c, d) from COMPAS$_{flip}$, (e, f) from UTICalc$_{LLM}$, and (g, h) from COMPAS$_{LLM}$. Steering with the vector learned from UTICalc$_{flip}$ at $\alpha = 75$ significantly reduced TE on both evaluation datasets, but it also significantly reduced accuracy on both. On UTICalc, TE decreased by 0.154 (95\% CI [$-0.160, -0.148$]) and accuracy decreased by 0.055 (95\% CI [$-0.086, -0.023$]). On COMPAS, TE decreased by 0.089 (95\% CI [$-0.093, -0.085$]) and accuracy decreased by 0.043 (95\% CI [$-0.054, -0.032$]). The vector learned from COMPAS$_{flip}$ behaved differently on the two evaluation datasets. On COMPAS, TE decreased modestly with increasing $\alpha$ up to approximately $\alpha = 40$, but surprisingly rose at larger strengths. At $\alpha = 75$, TE increased significantly by 0.165 (95\% CI [0.099, 0.230]), while accuracy decreased by $-0.025$ but was statistically insignificant (95\% CI [$-0.110, 0.060$]). On UTICalc, TE still significantly decreased at $\alpha=75$ by 0.125 (95\% CI [$-0.156, -0.097$]), and accuracy decreased by $-0.010$, but was statistically insignificant (95\% CI [$-0.040, 0.020$]). The vector learned from UTICalc$_{LLM}$ reduced TE on both evaluation datasets, but its effect on accuracy differed. On UTICalc, steering at $\alpha = 75$ significantly reduced TE by 0.021 (95\% CI [$-0.025$, $-0.017$]), while accuracy was exactly unchanged. The LLM-generated counterfactual set for UTICalc$_{LLM}$ is relatively small, and steering shifted predicted probabilities without flipping any predicted labels, so accuracy stayed constant while TE changed. On COMPAS, steering at $\alpha = 75$ significantly reduced TE by 0.018 (95\% CI [$-0.021$, $-0.016$]) and significantly reduced accuracy by 0.020 (95\% CI [$-0.032$, $-0.007$]). The vector learned from COMPAS$_{LLM}$ reduced TE on both evaluation datasets without a significant accuracy cost. On COMPAS, steering at $\alpha = 75$ significantly reduced TE by 0.046 (95\% CI [$-0.051$, $-0.041$]), and accuracy did not change significantly ($-0.006$, 95\% CI [$-0.015$, 0.003]). On UTICalc, steering at $\alpha = 75$ significantly reduced TE by 0.088 (95\% CI [$-0.104$, $-0.074$]), while accuracy was still exactly unchanged.

Across the four vectors, steering at $\alpha = 75$ significantly reduced TE in seven of eight vector--dataset settings, the exception being the COMPAS$_{flip}$ vector, where TE increased for the COPMAS data at larger $\alpha$ values. The vectors from LLM-derived counterfactuals produced smaller TE reductions (0.018 to 0.088) than vectors from race flip counterfactuals (0.089 to 0.154), but the cost to accuracy was of higher magnitude, especially using the UTICalc$_{flip}$ vector (\textasciitilde 5\% vs. \textasciitilde 1-2\%).

\section{Conclusion}
Steering vectors learned on one dataset reduced counterfactual unfairness on the other in most settings. TE fell significantly in seven of eight steering vector--dataset combinations. This suggests that layer-10 identified by activation patching captures a representation of race shared across tasks, so a vector learned to mitigate this bias works in both tasks. Transfer was not uniform, however. The COMPAS$_{flip}$ vector increased TE on COMPAS data at high steering strengths, and TE responded non-monotonically to $\alpha$ in several cases. Despite this, steering at $\alpha=75$ was surprisingly near optimal in all test cases. This provides another piece of evidence that large models may learn patterns or circuits across prediction tasks. 

The quality of the counterfactuals shaped the steering vector's efficacy as a bias mitigatior. Vectors from \textit{flip} counterfactuals produced the largest TE reductions (0.089 to 0.154) but also the largest accuracy costs. Additionally, we suspect the failure of mitigation on COMPAS in Figure \ref{fig:steer_mitigation}(d) relates to other features left invariant, but highly correlated with race. Vectors from LLM-generated counterfactuals produced smaller reductions (0.018 to 0.088) with little or no accuracy cost. A plausible explanation is that attribute flips can produce unrealistic inputs, so part of the learned direction reflects off-distribution changes and not race alone, whereas filtered counterfactuals yield a more targeted, and therefore smaller, measured effect. Our comparison does not perfectly isolate this effect, since the two methods also differ in sample size and filtering, and the filtered UTICalc$_{LLM}$ set is small (271 pairs). The results are consistent with counterfactual quality limiting how well a vector transfers bias mitigation across tasks, a hypothesis that future work could examine through more extensive empirical testing.

We optimized and evaluated steering on TE, and reductions were concentrated in the natural direct effect, with the indirect effect unchanged or slightly increased in several settings. Because TE, NDE, and NIE are all computed from the same factual--counterfactual pairs, the vector could in principle be learned to target any of these effects, for example, the NIE to address bias mediated through correlated features. We did not test this, and doing so is a natural extension of the method.

The results show that a steering vector learned from counterfactual pairs on one task can meaningfully reduce racial bias on another, which suggests that activation steering is a promising and inexpensive route to mitigating counterfactual unfairness without further training. The strength of this transfer depends on how the counterfactuals are constructed, with flip counterfactuals yielding larger reductions at some cost to accuracy and LLM-generated counterfactuals yielding smaller but cleaner ones. The gains also came mainly through the direct effect of race rather than effects mediated by correlated features. The consistency of transfer across settings indicates that the representation of race in large models may be shared across prediction tasks, and that steering offers a practical tool for both reducing and better understanding that bias.

\section*{Limitations}

In most cases, the steering strength we selected $(\alpha=75)$ was optimal, so our testing across these tasks was comparable. When steering strength may need to differ, it is less obvious whether the vectors mitigate the same bias. Because this value was not optimal in every setting (see Figure \ref{fig:steer_mitigation}(d)), a setting-specific $\alpha$ may be needed, making it less clear whether vectors mitigate the same bias at different strengths. Our study also covers one model across two tabular prediction tasks with a single binary protected attribute, and only mitigate TE. Future work is needed to test the framework's robustness across other models, model components, data modalities (e.g., image data), and sensitive attributes.


\section*{Generative AI Statement}

The authors used Claude Sonnet 5.5 to assist with table formatting, plot formatting, and grammar in writing the manuscript. The authors did not use generative AI to produce any methodology, experiments, or other key paper contributions. All authors also reviewed and edited the manuscript after any AI use. Counterfactual data used in the experiments were produced using a generative model. This is described in detail in the Methods.

\section*{Ethical Considerations}

Fairness interventions may be interpreted as guarantees. Our framework reduces counterfactual unfairness as measured by our metrics on the evaluated datasets and models, but it does not certify that a model is "fair" in general as it is an ongoing debate on how to define "fairness" generally. It should not be used to justify deploying a model in high-stakes settings such as criminal justice or healthcare without human, domain-specific validation, oversight, and review. 

The datasets reflect historical and societal biases. We use these datasets as benchmarks for fairness analysis only, and we do not make any ethical claims about the use of race as a predictor. Our methodology is intended to improve fairness in biased predictive models, not to justify or argue against the use of race or any sensitive feature in constructing new models.

The same techniques that identify and steer sensitive attribute in model activations could in principle be used to amplify bias rather than reduce it. Our evaluation is limited to a binary race variable, and we do not test other sensitive attributes. We have not tested other sensitive attributes or populations.

\bibliographystyle{ACM-Reference-Format}
\bibliography{refs}

\end{document}